\documentclass[11pt,a4paper]{article}
\usepackage[hyperref]{acl2021}
\usepackage{times}
\usepackage{latexsym}

\usepackage{enumitem}
\usepackage{xcolor,colortbl}

\usepackage[textsize=small,textwidth=0.60in]{todonotes}
\usepackage{microtype}

\usepackage{soul}
\usepackage{tikz}
\usetikzlibrary{calc}

\usepackage{amsmath}

\makeatletter
\newif\if@anonymize

\@anonymizetrue    

\if@anonymize
  \newcommand{\highlight@DoHighlight}{
    \fill [outer sep = -15pt, inner sep = 0pt, color=black]
          ($(begin highlight)+(0,8pt)$) rectangle ($(end highlight)+(0,-3pt)$) ;
  }

  \newcommand{\highlight@BeginHighlight}{
    \coordinate (begin highlight) at (0,0) ;
  }

  \newcommand{\highlight@EndHighlight}{
    \coordinate (end highlight) at (0,0) ;
  }

  \newdimen\highlight@previous
  \newdimen\highlight@current
  \newlength{\item@width}

  \DeclareRobustCommand*\anonymize{%
    \SOUL@setup
    \def\SOUL@preamble{%
      \begin{tikzpicture}[overlay, remember picture]
        \highlight@BeginHighlight
        \highlight@EndHighlight
      \end{tikzpicture}%
    }%
    \def\SOUL@postamble{%
      \begin{tikzpicture}[overlay, remember picture]
        \highlight@EndHighlight
        \highlight@DoHighlight
      \end{tikzpicture}%
    }%
    \def\SOUL@everyhyphen{%
      \discretionary{%
        \SOUL@setkern\SOUL@hyphkern
        \SOUL@sethyphenchar
        \tikz[overlay, remember picture] \highlight@EndHighlight ;%
      }{%
      }{%
        \SOUL@setkern\SOUL@charkern
      }%
    }%
    \def\SOUL@everyexhyphen##1{%
      \SOUL@setkern\SOUL@hyphkern
      \settowidth{\item@width}{##1}%
      \makebox[\item@width]{}%
      \discretionary{%
        \tikz[overlay, remember picture] \highlight@EndHighlight ;%
      }{%
      }{%
        \SOUL@setkern\SOUL@charkern
      }%
    }%
    \def\SOUL@everysyllable{%
      \begin{tikzpicture}[overlay, remember picture]
        \path let \p0 = (begin highlight), \p1 = (0,0) in \pgfextra
          \global\highlight@previous=\y0
          \global\highlight@current =\y1
        \endpgfextra (0,0) ;
        \ifdim\highlight@current < \highlight@previous
          \highlight@DoHighlight
          \highlight@BeginHighlight
        \fi
      \end{tikzpicture}%
      \settowidth{\item@width}{\the\SOUL@syllable}%
      \makebox[\item@width]{}%
      \tikz[overlay, remember picture] \highlight@EndHighlight ;%
    }%
    \SOUL@
  }
\else
  \newcommand{\anonymize}[1]{#1}
\fi
\makeatother

\aclfinalcopy 

\newcommand{\showRepoLink}

\title{HistBERT: A Pre-trained Language Model for Diachronic Lexical Semantic Analysis}

\author{Wenjun Qiu$^*$, Yang Xu$^{\dagger}$ \\
  $^*$Department of Electrical and Computer Engineering\\
  $^{\dagger}$ Department of Computer Science\\
  University of Toronto \\
  Toronto, Ontario, Canada \\
  \texttt{wenjun.qiu@mail.utoronto.ca}, \texttt{yangxu@cs.toronto.edu}
}

\date{}

\begin{document}
\maketitle
\begin{abstract}
Contextualized word embeddings have demonstrated state-of-the-art performance in various natural language processing tasks including those that concern historical semantic change. However, language models such as BERT was trained primarily on contemporary corpus data. To investigate whether training on  historical corpus data improves diachronic semantic analysis, we present a pre-trained BERT-based language model, HistBERT, trained on the balanced Corpus of Historical American English. We examine the effectiveness of our approach by comparing performance of the original BERT and that of HistBERT, and we report promising results in word similarity and semantic shift analysis. Our work suggests that the effectiveness of contextual embeddings in diachronic semantic analysis is dependent on the temporal profile of the input text and care should be taken in applying this methodology to study historical semantic change.

\end{abstract}

\section{Introduction}

Semantic change is an extensively studied topic in historical linguistics~\cite{traugott2001regularity}, but reliable quantification of semantic change remains a challenging problem~\cite{tahmasebi2018survey}. One of the most popular approaches to quantifying semantic change relies on distributional semantic models which have been applied to cases including but not restricted to the automatic detection, modeling, and evaluation of laws of semantic change~\cite{sagi2011tracing,cook2010automatically,Gulordava_2011,kulkarni2015statistically,kim_temporal_2014, krol_past_2017, rosenfeld_deep_2018, pmlr-v70-bamler17a,schlechtweg2020semeval,xu_Kemp_2015, hamilton_diachronic_2016, dubossarsky_outta_2017}. 


Recent development in natural language processing (NLP) shows that  contextualized embeddings outperform word embeddings in quantifying word meaning and semantic change~\cite{peters_deep_2018, elmo}. For instance, BERT has achieved state-of-the-art performance on a number of natural language understanding tasks~\cite{devlin_bert:_2019}. Building on these studies, ~\citealt{giulianelli_analysing_2020} and~\citealt{hu_diachronic_Ecological} construct contextualized embeddings for words in a diachronic setting and investigate word meaning change in vector space over historical times. Their work suggests that contextualized embeddings offer an advantage in the quantification and detection of historical semantic change.  

While BERT-based models have shown promising results, we challenge one of the basic assumptions in applying pre-trained BERT models for diachronic semantic analysis: BERT was typically trained on BooksCorpus~\cite{zhu_google_book} and English Wikipedia\footnote{\url{https://www.english-corpora.org/wiki/}} which often do not appropriately reflect language use in historical times. Although the corpora for pre-training BERT may contain some text from  history, modern corpora including Wikipedia contain mostly contemporary English text. This skewed pre-training input leads naturally to several  research questions that motivate our study: 
\begin{enumerate}
  \item Is it valid to use pre-trained BERT to model historical semantic change?
  \item Does this approach lead to semantic representations that are biased toward modern language use? 
  \item Can additional training on (balanced) historical data help to improve the precision of BERT in quantifying historical semantic change?
\end{enumerate}

These research questions are partly inspired by the study of~\citealt{giulianelli_analysing_2020}, where it was observed that issues arise when the original pre-trained BERT model was applied directly to quantifying semantic change (without controlling for the temporal profile of the input text). One prominent example described in \citealt{giulianelli_analysing_2020} was the insensitivity of BERT in teasing apart historical and modern usages of the English word ``coach''. Specifically, the pre-trained BERT model considered that the similarity between \emph{stage coach} and \emph{go coach} to be less than the similarity between \emph{stage coach} and \emph{Cinderella's coach}. We believe that such problematic cases exist precisely due to the lack of care in training BERT on  historical text, which forbade the model to disambiguate nuanced usage differences across historical time periods. 

In order to investigate the validity of this claim, we present a simple alternative approach dubbed HistBERT, a BERT-based language model that assimilates high-quality and more balanced historical data as opposed to relying on training data skewed toward the contemporary period. We pre-train HistBERT on longitudinal text data, particularly the Corpus of Historical American English (COHA)~\cite{davies_expanding_2012}, which is a well-known historical corpus containing fictions and non-fictions, newspapers, and magazines from 1820s to 2010s. We demonstrate the effectiveness of HistBERT by comparing its precision on historical semantic similarities against the original BERT model, and we make the following contributions: 

\begin{enumerate}
  \item We construct contextualized word embeddings by leveraging unsupervised pre-training on BERT using the COHA dataset. The pre-trained language model, HistBERT, is openly available here: \url{https://drive.google.com/drive/folders/1E4jZPzCatmS73-Mg7iacq4SXROSPsC6b?usp=sharing}.
  
  \item We investigate the effectiveness of our approach by comparing performance of the original BERT and that of HistBERT. We perform evaluation on various semantic analysis tasks, and observe promising results in word similarity and semantic shift analysis. The source code is available at: \url{https://github.com/wendyqiu/diachronicBert}.
\end{enumerate}

The remaining part of this paper is orgainzed as follows. We review related work in Section~\ref{related_work}. We  describe the  details of our model design and implementation for HistBERT in Section~\ref{design}. We present the results in Section~\ref{sec:evaluation}, along with with model interpretation and discussion. We close by addressing limitations and future directions in Section~\ref{conclusion}.

\section{Related Work} \label{related_work}

\subsection{Diachronic Semantic Modeling}

Pre-trained language model has been shown to be effective in lexical semantic change analysis~\cite{tahmasebi_survey_2019}. Existing models can be categorized into two major types, namely the form-based and sense-based approach~\cite{Kutuzov_2018}. The difference lies in that form-based models compute an embedding for each word, whereas sense-based models encode each word sense into a vector. As an example,~\citealt{hamilton_diachronic_2018} constructed HistWords, the openly available diachronic embeddings generated from PPMI, SVD, and SGND (word2vec). More recent methods presented probabilistic language model for time-stamped text data to track the semantic evolution of individual words~\cite{pmlr-v70-bamler17a,rudolph_dynamic_2018}. Despite the fact that these approaches have achieved reasonable performance, form-based approaches aggregate all senses of a word into a single representation~\cite{traugott_principles_1991} which may not be desirable in the case of modeling historical semantic shifts. 

To overcome  such an issue, existing studies have explored the sense-based approach.~\citealt{frermann_bayesian_2016} proposed a Bayesian-based model named SCAN to capture how a word's senses evolve over time. In their model, senses are expressed as a probability distribution over words, which can also change over time. However, such models assume that time is discrete, divided into consecutive intervals. 

Recent studies have investigated the potential of contextualized word representations to analyzing lexical semantic change. \citealt{hu-etal-2019-diachronic} proposed a sense representation and tracking framework based on BERT and achieved a substantial improvement in word meaning change detection. Using similar contextualized word embeddings,~\citealt{giulianelli_analysing_2020} proposed a model without lexicographic supervision and demonstrated that the detected semantic shifts are positively correlated with human judgements. They proposed different metrics to quantify lexical semantic changes in addition to the contextualized word representations, and demonstrated a positive correlation between model predictions and human judgements. In this study, we focus on extending this recent work and pre-train the contextualized language model with more balanced historical text---an important aspect that has not been explored previously. 

\subsection{Pre-trained Contextualized Language Models}

Pre-trained contextualized lanaguage models such as BERT leverage the transfer learning technique to achieve good performance in various downstream NLP tasks~\cite{devlin_bert:_2019}. However, as it was trained on general domain corpora such as Wikipedia, it lacks ability to perform specialized tasks in certain niche areas. Existing work has suggested how to improve the basic BERT by further training on domain-specific corpora. These embeddings were proven to be more effective than the original general-purpose BERT embeddings. For example, SciBERT~\cite{beltagy_scibert:_2019}, which targets specific on scientific articles, and BioBERT~\cite{lee_biobert:_2019}, which is constructed for biomedical text mining.

In this work, we identify the potential bias in BERT-based diachronic semantic analysis due to relying on modern word usages as input and a lack of historical input text, and we show how altering the pre-training input makes a difference in the quantification of historical word semantics.

\section{Methodology} \label{design}
Here we describe the technical details of how we constructed HistBERT. Background information on the training corpus is covered in Section~\ref{method:corpus}. The design process of HistBERT is divided into three steps: pre-processing  (Section~\ref{method:preprocess}), pre-training (Section~\ref{method:train}) and representation (or feature) extraction (Section~\ref{method:extract}). We build upon the open-sourced BERT\footnote{\url{https://github.com/google-research/bert}} and follow the official guidelines. Note that our implementation of the BERT-based models slightly differs from that of~\citealt{giulianelli_analysing_2020}, which is developed using \textit{Hugging Face}'s implementation. For computational efficiency, HistBERT is constructed using \emph{bert-base-uncased}, which consists of 12 hidden layers with 768 dimensions.

\subsection{Training Corpus} \label{method:corpus}
For the purpose of diachronic semantic analysis, we  perform further pre-training additionally to the original BERT using a large corpus with balanced historical text. This way, the model would be exposed to language usages throughout the historical period in question, and the updated model should  in principle capture contextualized meanings that are not presented in modern English. In this study, we use the COHA corpus, which is one of the largest structured corpora of historical English to date. COHA contains more than 475M words, extracted from various genres, including fictions, non-fictions, newspapers, magazines and so on. The historical text ranges from the 1820s to the 2010s~\cite{davies_expanding_2012}. To ensure the amount and quality of our training data, we focus on the 10 decades between the 1910s and the 2000s, where there are more than 20M texts for each decade.

\subsection{Pre-processing} \label{method:preprocess}
To pre-process the COHA dataset, we use the default BERT tokenizier and perform tokenization on sentence level. All textual data are normalized by lower-casing the input, stripping out accent markers, splitting punctuation, and handling abbreviations embedded in the dataset.  We perform an additional step to eliminate the mismatch between the tokenization schemes of BERT and COHA. For example, \emph{$\langle do \rangle$, $\langle n't \rangle$} for COHA, and \emph{$\langle don't \rangle$} for BERT. We remove such contractions and converted them back to the raw-looking form. As suggested in~\citealt{Gulordava_2011}, lemmatization has little effect on BERT models trained on English texts. Therefore, we do not lemmatize our pre-training texts. Finally, we organize our input files according to BERT's pre-training requirements: 
\begin{enumerate} 
  \item One sentence per line, which would be truncated during training if exceeds \emph{max\_seq\_length}.
  \item Documents are separated by empty newline.
  \item Truncate 2\% of input segments randomly to improve the model robustness to non-sentential input. 
\end{enumerate}

\subsection{Pre-training} \label{method:train}

HistBERT models are constructed by continuing pre-training the original BERT. For each model, the training starts from the last checkpoint of the original BERT. Models are trained on the following two tasks: \emph{masked LM} and \emph{next sentence prediction}, which align with the original BERT model. We use the configuration shown in Table~\ref{table:hyperparameters} to train our models on the openly available BERT source code. We prepare multiple models trained on different settings to examine the performance and robustness of HistBERT performs under different situations. 

\begin{table*}[h!]
  \centering
\begin{tabular}{|l|c|c|c|c|}
\hline
\textbf{Hyperparameters}            & \multicolumn{1}{l|}{\textbf{HistBERT-proto}} & \multicolumn{1}{l|}{\textbf{HistBERT-5}} & \multicolumn{1}{l|}{\textbf{HistBERT-10}} \\ \hline
\textit{max\_seq\_length}           & 128                                               & 128                                                         & 128                                                                                                        \\ \hline
\textit{max\_predictions\_per\_seq} & 20                                                & 20                                                          & 20                                                                                                               \\ \hline
\textit{masked\_lm\_prob}           & 0.15                                              & 0.15                                                        & 0.15                                                                                                          \\ \hline
\textit{train\_batch\_size}         & 32                                                & 32                                                          & 32                                                                                                               \\ \hline
\textit{num\_warmup\_steps}         & 10,000                                            & 500,000                                                     & 500,000                                                                                            \\ \hline
\textit{num\_train\_steps}          & 10,000                                            & 200,000                                                     & 200,000                                                                                             \\ \hline
\textit{learning\_rate}             & 2e-5                                              & 2e-5                                                        & 2e-5                                                                                                    \\ \hline
\end{tabular}
 \captionsetup{justification=centering}
 \caption{Pre-training hyperparameters for different HistBERT models}
 \label{table:hyperparameters}
 
\end{table*}

We train a prototype model, \textbf{HistBERT-proto}, to verify the feasibility of our proposed idea, and to present results for a specific use case (Section~\ref{sec:pca}). The prototype model is trained on selected data, with reduced the breadth and depth. As mentioned, we focus on the 20th century, where there are more than 20M texts per decade in COHA. For \textbf{HistBERT-proto}, we further reduce the depth of training data by selecting documents that cover a specific keyword, in this case, \emph{``coach''}. We select this keyword as it is identified as a problematic word in~\citealt{giulianelli_analysing_2020}. Based on the original BERT model, the similarity between \emph{stage coach} and \emph{go coach} is smaller than the similarity between \emph{stage coach} and \emph{Cinderella's coach}, which contradicts with the common judgements. We suspect the reason to be the lack of training on historical text data in the original BERT model. To test our hypothesis, we build a prototype using textual data that contain the keyword ``coach''. The training hyperparameters for this prototype is much smaller than other models. We limit the number of training steps to prevent overfitting, because the training set consists of only 280M texts, much smaller than the full COHA. 

A richer version of the  model, \textbf{HistBERT-5}, is also trained on partial data selected from COHA, covering only 5 decades of texts, ranging from 1910s to 1950s. In other words, it is to expose sufficient historical data to the original BERT model, but focus only on the first half decade.

The most comprehensive model, \textbf{HistBERT-10}, is trained on data selected from 1910s to 2000s (i.e., 10 decades). This model is trained on a much larger subset of COHA. We eliminate texts from the last decade (2010s), as they may overlap with the original training data of BERT. The 10-decade textual data also align with the time period of our evaluation set DUPS which was used in \citealt{giulianelli_analysing_2020}. We investigate the impact of different training data on the model performance by comparing these HistBERT models. Further analysis is covered in~\ref{sec:evaluation}. 

Due to the large amount of data, we performed additional training using a larger \emph{max\_seq\_length}, allowing the model to further learn its positional embeddings. The reason why we did not use \emph{max\_seq\_length}$=512$ for all its training iterations is that longer sequences are disproportionately expensive due to the attention nature of BERT. As suggested by the original authors\footnote{\url{https://github.com/google-research/bert}}, a batch of 32 sequences of length 512 is much more expensive than a batch of 128 sequences of length 128, despite having the same amount of training data. 

All 3 versions of HistBERT are trained from the last checkpoint of \emph{bert-base-uncased}. The pre-training process is done on a single Google Cloud v3-8 TPU with 64 GB memory. The virtual instance was on 2 Intel Haswell vCPUs with 7.5 GB memory, and a 250 G of hard drive is associated with each instance. The output models were stored in Google Cloud Storage Bucket. Each finalized model has an average size of 1.3 GB. Each model takes up to 34 hours to complete.

\subsection{Representation Extraction} \label{method:extract}
Using different versions of HistBERT, we construct data representations from the pre-trained language models. We extract word embeddings from HistBERT using a similar approach as the prior work~\cite{giulianelli_analysing_2020}: for each focus word $w$, we will find all sentences that contain $w$, with its position denoted as $i$. Next, we feed these sentences into the pre-trained model and obtain the according sentence embeddings. To find the exact embedding for each focus word, we extract embeddings from position $i$ in each hidden layers and sum them dimension-wise. For efficiency, we focus on the last 4 hidden layers, which is the suggested number of layers to be used in BERT's feature vector extraction.

To measure word sense similarity, we compute the distance between the selected vector representations. To quantify  lexical semantic change, we compute a cosine similarity score for each pairwise comparison between two word usages from different times. We next describe the details for model evaluation.

\section{Evaluation and Results} \label{sec:evaluation}

We evaluate HistBERT by comparing its performance against the original BERT model. We observe an overall improvement in the positively correlation with human annotations (Section~\ref{dups}). We further perform a case study on the specific keyword ``coach'', by providing a low-dimensional visualization in Section~\ref{sec:pca}, and analyzing the embedding shifts in Section~\ref{cosine}. The experiments were conducted on a local Ubuntu 20.04 system, with a single Nvidia 1070 GPU and an Intel i7-7700 CPU @ 3.60GHz with 16.0 GB of memory.

\subsection{Evaluation against DUPS Human Similarity Data} \label{dups}

\begin{table*}[ht]
\centering
\begin{tabular}{|r|ll|ll|ll|ll|}
\hline
\multicolumn{1}{|c|}{} & \multicolumn{2}{c|}{\textbf{BERT}}                                        & \multicolumn{2}{c|}{\textbf{HistBERT-proto}}                              & \multicolumn{2}{c|}{\textbf{HistBERT-5}}                                  & \multicolumn{2}{c|}{\textbf{HistBERT-10}}                                 \\ \cline{2-9} 
\multicolumn{1}{|l|}{} & \multicolumn{1}{c|}{\textbf{$\rho$}}       & \multicolumn{1}{c|}{\textbf{p}} & \multicolumn{1}{c|}{\textbf{$\rho$}}       & \multicolumn{1}{c|}{\textbf{p}} & \multicolumn{1}{c|}{\textbf{$\rho$}}       & \multicolumn{1}{c|}{\textbf{p}} & \multicolumn{1}{c|}{\textbf{$\rho$}}       & \multicolumn{1}{c|}{\textbf{p}} \\ \hline
\textbf{leaf}          & \cellcolor[HTML]{DDEBF7}0.2405          & \cellcolor[HTML]{DDEBF7}0.0010  & \cellcolor[HTML]{FCE4D6}\textbf{0.3059} & \cellcolor[HTML]{FCE4D6}0.0010  & \cellcolor[HTML]{E2EFDA}\textbf{0.2896} & \cellcolor[HTML]{E2EFDA}0.0010  & \cellcolor[HTML]{FFF2CC}\textbf{0.2742} & \cellcolor[HTML]{FFF2CC}0.0010  \\
sphere                 & \cellcolor[HTML]{DDEBF7}\textbf{0.4961} & \cellcolor[HTML]{DDEBF7}0.0010  & \cellcolor[HTML]{FCE4D6}0.4339          & \cellcolor[HTML]{FCE4D6}0.0050  & \cellcolor[HTML]{E2EFDA}0.4359          & \cellcolor[HTML]{E2EFDA}0.0070  & \cellcolor[HTML]{FFF2CC}0.4124          & \cellcolor[HTML]{FFF2CC}0.0160  \\
\textbf{coach}         & \cellcolor[HTML]{DDEBF7}0.4338          & \cellcolor[HTML]{DDEBF7}0.0010  & \cellcolor[HTML]{FCE4D6}\textbf{0.4443} & \cellcolor[HTML]{FCE4D6}0.0020  & \cellcolor[HTML]{E2EFDA}0.3625          & \cellcolor[HTML]{E2EFDA}0.0090  & \cellcolor[HTML]{FFF2CC}0.3931          & \cellcolor[HTML]{FFF2CC}0.0080  \\
net                    & \cellcolor[HTML]{DDEBF7}\textbf{0.2816} & \cellcolor[HTML]{DDEBF7}0.0010  & \cellcolor[HTML]{FCE4D6}0.2201          & \cellcolor[HTML]{FCE4D6}0.0010  & \cellcolor[HTML]{E2EFDA}0.2051          & \cellcolor[HTML]{E2EFDA}0.0020  & \cellcolor[HTML]{FFF2CC}0.2182          & \cellcolor[HTML]{FFF2CC}0.0020  \\
\textbf{federal}       & \cellcolor[HTML]{DDEBF7}0.1227          & \cellcolor[HTML]{DDEBF7}0.0010  & \cellcolor[HTML]{FCE4D6}\textbf{0.1315} & \cellcolor[HTML]{FCE4D6}0.0010  & \cellcolor[HTML]{E2EFDA}\textbf{0.1596} & \cellcolor[HTML]{E2EFDA}0.0010  & \cellcolor[HTML]{FFF2CC}\textbf{0.1514} & \cellcolor[HTML]{FFF2CC}0.0010  \\
\textbf{compact}       & \cellcolor[HTML]{DDEBF7}0.2033          & \cellcolor[HTML]{DDEBF7}0.0080  & \cellcolor[HTML]{FCE4D6}\textbf{0.2702} & \cellcolor[HTML]{FCE4D6}0.0010  & \cellcolor[HTML]{E2EFDA}\textbf{0.2599} & \cellcolor[HTML]{E2EFDA}0.0010  & \cellcolor[HTML]{FFF2CC}\textbf{0.2708} & \cellcolor[HTML]{FFF2CC}0.0010  \\
\textbf{signal}        & \cellcolor[HTML]{DDEBF7}0.2241          & \cellcolor[HTML]{DDEBF7}0.0090  & \cellcolor[HTML]{FCE4D6}\textbf{0.2768} & \cellcolor[HTML]{FCE4D6}0.0010  & 0.1583          & 0.0830  & 0.1633                                  & 0.0850                          \\
\textbf{mirror}        & \cellcolor[HTML]{DDEBF7}0.4985          & \cellcolor[HTML]{DDEBF7}0.0100  & \cellcolor[HTML]{FCE4D6}0.4945          & \cellcolor[HTML]{FCE4D6}0.0100  & \cellcolor[HTML]{E2EFDA}\textbf{0.5307} & \cellcolor[HTML]{E2EFDA}0.0060  & \cellcolor[HTML]{FFF2CC}\textbf{0.5629} & \cellcolor[HTML]{FFF2CC}0.0030  \\
\textbf{optical}       & \cellcolor[HTML]{DDEBF7}0.1902          & \cellcolor[HTML]{DDEBF7}0.0220  & \cellcolor[HTML]{FCE4D6}\textbf{0.2542} & \cellcolor[HTML]{FCE4D6}0.0020  & \cellcolor[HTML]{E2EFDA}\textbf{0.2739} & \cellcolor[HTML]{E2EFDA}0.0010  & \cellcolor[HTML]{FFF2CC}\textbf{0.2554} & \cellcolor[HTML]{FFF2CC}0.0030  \\
spine         & \cellcolor[HTML]{DDEBF7}\textbf{0.1819}          & \cellcolor[HTML]{DDEBF7}0.0380  & 0.1593                                  & 0.0970                          & 0.1534                                  & 0.1470                          & 0.1502                                  & 0.1490                          \\
\textbf{card}          & 0.3987                                  & 0.0535                          & \cellcolor[HTML]{FCE4D6}\textbf{0.4987} & \cellcolor[HTML]{FCE4D6}0.0160  & \cellcolor[HTML]{E2EFDA}\textbf{0.5209} & \cellcolor[HTML]{E2EFDA}0.0090  & \cellcolor[HTML]{FFF2CC}\textbf{0.5205} & \cellcolor[HTML]{FFF2CC}0.0100  \\
\textbf{brick}         & 0.4462                                  & 0.2340                          & \cellcolor[HTML]{FCE4D6}0.4235          & \cellcolor[HTML]{FCE4D6}0.0400  & \cellcolor[HTML]{E2EFDA}\textbf{0.5012} & \cellcolor[HTML]{E2EFDA}0.0140  & \cellcolor[HTML]{FFF2CC}\textbf{0.5376} & \cellcolor[HTML]{FFF2CC}0.0030  \\
\textbf{virus}         & 0.1968                                  & 0.3870                          & \cellcolor[HTML]{FCE4D6}\textbf{0.5618} & \cellcolor[HTML]{FCE4D6}0.0040  & \cellcolor[HTML]{E2EFDA}\textbf{0.4214} & \cellcolor[HTML]{E2EFDA}0.0480  & \cellcolor[HTML]{FFF2CC}\textbf{0.4684} & \cellcolor[HTML]{FFF2CC}0.0470  \\ \hline
\textbf{disk}          & 0.1066                                  & 0.5010                          & 0.1735                                  & \textbf{0.2710}                 & 0.1337                                  & \textbf{0.3860}                 & -0.0233                                 & 0.8790                          \\
\textbf{energy}        & -0.0151                                 & 0.9260                          & 0.1487                                  & \textbf{0.3820}                 & 0.2591                                  & \textbf{0.1260}                 & 0.2209                                  & \textbf{0.2360}                 \\
\textbf{virtual}       & 0.0088                                  & 0.9580                          & -0.0415                                 & \textbf{0.7790}                 & 0.0547                                  & \textbf{0.6960}                 & 0.0878                                  & \textbf{0.5050}                 \\ \hline
\end{tabular}
 \captionsetup{}
 \caption{Spearman's rank correlations of different BERT models on DUPS. Words that have a significant positive correlation are highlighted. Results that outperform the \textbf{BERT} baseline are marked in bold: in the cases where the result is significant, the coefficient $\rho$ will be marked; vice-versa, the p-value will be marked. A line is shown to separates the 3 words that do not pass the test of significance in any models.}
 \label{table:dups_eval}
\end{table*}

As an overall evaluation, we determine whether the word embeddings output from HistBERT align with human similarity judgements, and how well it performs in comparison to the original BERT model. In this experiment, we test our models using the Diachronic Usage Pair Similarity (DUPS) dataset\footnote{\url{https://doi.org/10.5281/zenodo.3773250}} created by~\citealt{giulianelli_analysing_2020}. 

The DUPS dataset contains crowdsourced similarity judgements of English word usage pairs from different time decades between the 20th century, each being labelled by five human annotators. They sampled 16 words uniformly at random from the GEMS dataset~\cite{Gulordava_2011}, resulting in 3,285 usage pairs. Each DUPS usage pair contains two text snippets that contains the same focus word. Each snippet contains a label of time interval to which the usage sentence belongs. A sample usage pair for the keyword ``virus'' is shown as follows:

\begin{quote}
\textit{Solomon's software, recently acquired by data-fellows, offers a pair of virus scanners: Dr.Solomon's Anti-\underline{virus} Deluxe for Windows 95 and Dr.Solomon's Virex for the macintosh OS.} \textbf{[1990--2000]}
\end{quote}

\noindent and 

\begin{quote}

\textit{\underline{Virus}, of many forms, appears in every cancer patient and vitiates his blood, upsets the biochemistry homologous to the normal for the species.} \textbf{[1910--1920]}
\end{quote}

By averaging the five scores, we obtain a single human similarity judgment for each usage pair. We create a dictionary of matrices to record the average scores for all usage pairs, denoted as the \emph{matrices of DUPS's similarity scores}. 

We use the following steps to compute similarity scores using the original BERT and our HistBERT models: given the sentence snippets provided in DUPS, we obtain sentence embeddings from each BERT-based model, and extract the word embedding for the focus word, resulting in two contextualized word vectors for each usage pair. We compute the similarity scores between the two usages. For each model, we create a dictionary of matrices that have the same dimension as the \emph{matrices of DUPS's similarity scores}. We compute the Spearman's rank correlation between the human similarity scores and the similarity scores obtained from each model. We use the Scikit-bio Python library\footnote{\url{http://scikit-bio.org/docs/0.1.3/index.html}}, which is developed based on the Mantel test~\cite{Mantel209}. Table~\ref{table:dups_eval} presents the correlation coefficients $\rho$ and p-values obtained for each focus word. 

Recall that we use the tensorflow-based BERT library to extract features, and create word embeddings based on the last 4 layers only. Because of the set up difference between our work and the prior work by~\citealt{giulianelli_analysing_2020}, the correlation results of BERT model we present are slightly different from what they reported. However, the words with significant positive correlations remain the same, summing to 10 out of 16 words, as highlighted in the \textbf{BERT} column in Table~\ref{table:dups_eval}. 

We observe an improvement in both the correlation coefficients and p-values. Firstly, the number of words with significant positive correlation (p $\leq 0.05$) increases from 10 out of 16 to 12, 11, and 12 respectively for \textbf{HistBERT-proto}, \textbf{HistBERT-5} and \textbf{HistBERT-10}. Compared with \textbf{BERT}, our best model, \textbf{HistBERT-10}, shows an increase in the average correlation of the significant words, from $0.31$ to $0.35$. The range also increases from $[0.12, 0.5]$ to $[0.15, 0.56]$. The performance difference can be explained by the choices of training data: \textbf{HistBERT-10} is trained with the largest amount of additional textual data among all models. At the same time, it covers the full 10 decades from the 1910s to 2000s, which align with the target time period of DUPS. 

For the 13 words that have any significant positive correlations in one or more models, we observe 10 words that show an improvement under the construction of HistBERT. For the remaining 3 words, as shown in the bottom of the table below the line, HistBERT models result in lower p-values, indicating a stronger deviation against the null hypothesis. The gain of significance is likely due to the increased size of training samples for senses that occur more frequently during the 20th century, yet are not very popular in modern English. Among all 16 words, there is only 1 word showing a deviation from significance. We believe that introducing more historical usages may decrease the prediction accuracy on novel senses, as the language model shifts its focus towards the historical usages since they are used in the most recent training process. This may be related to catastrophic interference, which is the tendency of a neural network to forget previously learned information upon learning new information~\cite{Ratcliff_Roger_1990, mccloskey_catastrophic_1989}. 

Overall, we observe an improvement in 13 out of 16 words. The remaining 3 words are ``sphere'', ``net'' and ``spine''. We believe that the different senses of these words do not specifically related to historical texts in the 20th century. In such situations, there is no obvious advantage of the HistBERT models over the original BERT, despite being trained on  additional COHA data.

\subsection{Case Study and Visualization of Semantic Shifts} \label{sec:pca}

\begin{figure*} [ht]
    \centering
    \includegraphics[width=2.0\columnwidth]{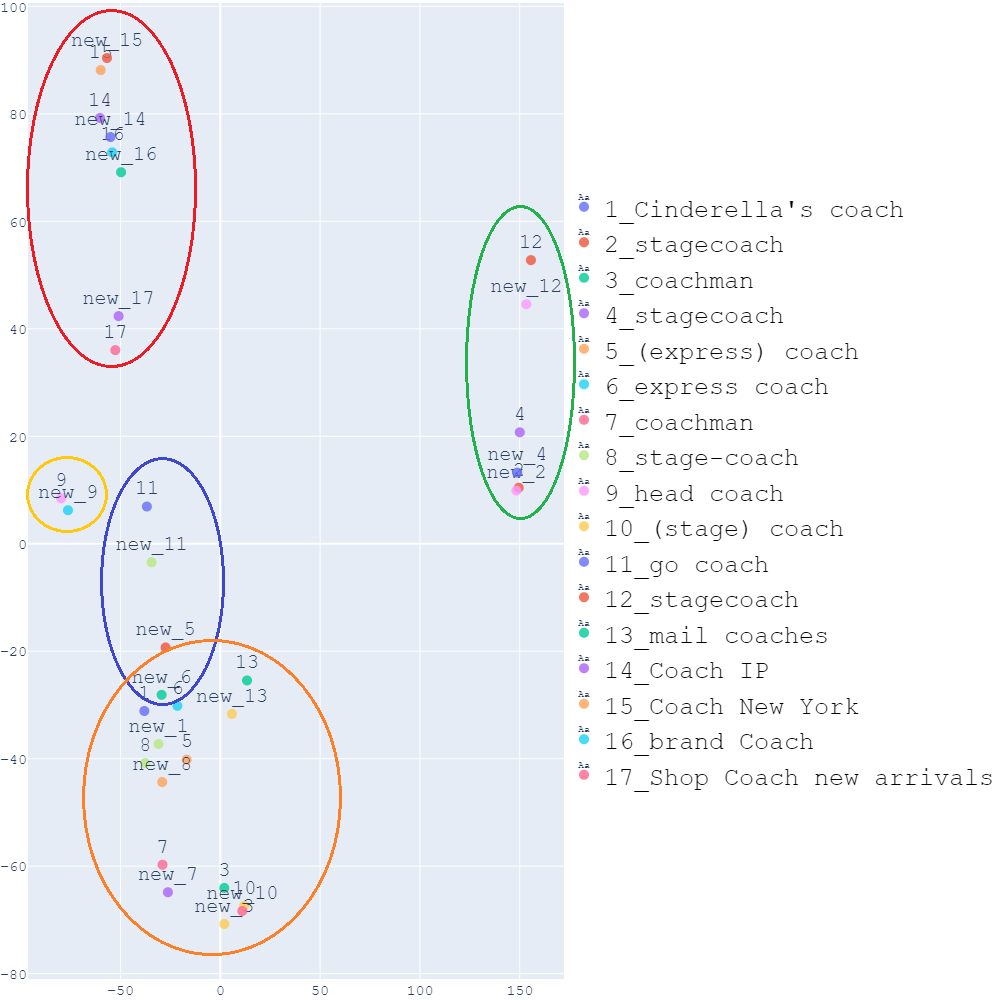}
    \caption{PCA visualization for usages of ``coach'' before and after training on historical data: vectors of \textbf{BERT} are labelled with numbers only, whereas vectors produced by \textbf{HistBERT-proto} are labelled with [number]\_new.}
    \label{fig:pca_overall}
\end{figure*}

To gain insights into the HistBERT model, we perform a case study to further illustrate how time-relevant data improves performance in diachronic semantic analysis using BERT. 

We focus on the keyword ``coach'' and visualize the embedding shifts from \textbf{BERT} and \textbf{HistBERT-proto}. We prepare 12 sentences that cover the focus word ``coach''. All sentences are selected from either COHA or online resources such as Wikipedia. Because the focus word occurs more than once in some sentences, the total number of usages generated from these sentences is 17, as shown in the legend of Figure~\ref{fig:pca_overall}. These sample sentences cover various usages of the word ``coach'' grouped into 4 major categories: 
\begin{enumerate}  
  \item An enclosed vehicle with four wheels, pulled by horses, and in which people used to travel.
  \item A large, comfortable bus that carries passengers on long journeys.
  \item Someone who trains a person or team of people in a particular sport.
  \item An American luxury design house specializing in handbags, luggage and ready-to-wear.
\end{enumerate}

As seen in the visualization, the original usages of ``coach'' generated from BERT are labelled with numbers, and the same usages created by \textbf{HistBERT-proto} are labelled as \emph{[number]\_new}. Note that sample 9 (circled in yellow) remains at the same position before and after, whereas other types of usages shift away from their original positions and where sample 9 locates. This is expected because the sense of a sport coach is very different from the remaining categories. We used sample 9 as a stable reference (or anchors) to illustrate that the shifts of other usages are not random.

The right-most cluster is separated from the remaining usages (circled in green). We observe that sample 4 and 12 move towards sample 2, forming a closer cluster. The reduction of intra-cluster distance indicates that after training on historical data, \textbf{HistBERT-proto} is able to better differentiate \emph{category 1} from other usage types: all three samples in this cluster refer to stagecoaches. 

In the cluster circled in blue, sample 5, 6, and 11 all refer to \emph{category 2}, namely ``express coach'', ``coach travel'', and ``go coach''. After the additional training on COHA, our language model produces a tighter cluster of these samples, which share similar meanings. Also notice that sample 1, which refers to Cinderella's pumpkin coach, moves away from the ``express coach'' cluster.  

Observe that sample 1 not only shifts away from the cluster of \emph{category 2}, it also moves towards sample 8, and to the further bottom where sample 7, 3 and 10 locate. These samples refer to usages of ``stage coach'', ``stage-coach'', and ``coachman''. Despite sharing a similar sense, they are grouped into a different cluster from sample 2, 4 and 12, due to the difference in their lexical usages. The movement of sample 1 is also desirable: as an existing issue of using BERT directly to generate usage representations, the usages of ``go coach'' and ``stage coach'' are mixed into one cluster, whereas ``Cinderella's coach'' is left out in a separated cluster. In our newly constructed model, we observe that the ``go coach'' cluster is moving away from the ``stage coach'' cluster, and ``Cinderella's coach'' is moving towards ``stage coach''. 

Also, observe that sample 5, which refers to ``express coach'', shifts away from the middle cluster that contains ``Cinderella's coach'' and ``stage-coach'', to the very top where ``express coach'' and ``go coach'' samples are.

The remaining samples are located on the top-left corner. Sample 14, 15, 16 and 17 (circled in red) belong to \emph{category 4}: the American brand \emph{Coach New York}. Observe that sample 14, 16, and 17 are moving towards each others and form a tighter cluster, reducing the intra-cluster distance; sample 15 shifts along the x-axis of the cluster’s direction, but in opposite y-axis direction.  

To summarize, we observe a decrease in the intra-cluster distance, whereas the inter-cluster distance increases. We also observe some substantial shifts that correct the misunderstood word sense representations from the original BERT model. 

\subsection{Quantification of Semantic Shifts in Embedding Space} \label{cosine}

To go beyond the qualitative interpretations, we also quantify the shifts in contextualized word embeddings presented in the PCA visualization. We measure the pairwise similarity between each usage pair and compare the difference between what \textbf{BERT} and \textbf{HistBERT-proto} produce. We define this quantity as the inverse of \textit{embedding shift} due to the additional training on historical data. 

The average embedding shift of all 17 instances of ``coach'' is $-0.012$, which is closed to $0$, indicating that the overall shift remains balanced. The maximum increase of cosine similarity indicates that the usage pair undergoes a change and become the most similar. We observe that the pair of ``Cinderella's Coach'' and ``stagecoach'' has the maximum increase in similarity ($0.066$). On the other hand, the maximum decrease occurs in usage pairs that become the least similar. In our case, ``(express) coach'' and ``(stage) coach'' has a negative change in similarity: $-0.13$. 

Both of these results confirm our hypothesis that additional training on historical data correct the misunderstood word sense representations from the original model, and helps BERT to better represent words that undergo semantic shifts. The effectiveness of contextual embeddings in diachronic semantic analysis is dependent on the input quality and specifically temporal profile of the input text.

\section{Conclusion} \label{conclusion}

We present HistBERT, a pre-trained BERT-based language model for historical semantic analysis. We construct various versions of the model and evaluate them against human similarity judgments. We also present a visualization to illustrate semantic shifts due to the additional training on historical textual data. 

Our results indicate that training on historical data is beneficial for diachronic semantic analysis using BERT. We suggest that the temporal profile of the textual input should be carefully controlled for in order to optimize the precision of the contextualized language model in a diachronic setting. 

Future work could apply HistBERT to detect semantic shifts at a systematic scale. For instance, experimenting on word usage clustering and changes similar to \citealt{giulianelli_analysing_2020}, or  detecting lexical semantic changes using unsupervised methods, as described in SemEval-2020 Task 1 \cite{schlechtweg_semeval-2020_2020}, are promising venues for further exploration of the methodology that we have presented here.

\section{Acknowledgements}

We thank David Lie for helpful feedback on this manuscript.

\bibliographystyle{acl_natbib}
\bibliography{anthology,acl2021}

\end{document}